\documentclass{article}
\usepackage[final]{corl_2026}
\usepackage{amsmath}
\usepackage{amssymb}
\usepackage{graphicx}
\usepackage{enumitem}
\usepackage{multirow}
\usepackage{booktabs}
\usepackage{wrapfig}
\usepackage{placeins}
\usepackage{needspace}
\usepackage{afterpage}
\usepackage{capt-of}
\usepackage{subcaption}
\title{Anchoring Attention in Clutter: Mask-Guided Latent Distillation for Vision-Language-Action Towards Dexterous Grasping}

\author{
  \textbf{Xuanyu Zhang}\textsuperscript{\normalfont 1,2} \quad
  \textbf{He Li}\textsuperscript{\normalfont 1,2} \quad
  \textbf{Xinxiang Lou}\textsuperscript{\normalfont 1,2} \quad
  \textbf{Lu Chen}\textsuperscript{\normalfont 2} \\
  \textbf{Yuxiang Cui}\textsuperscript{\normalfont 2} \quad
  \textbf{Zhongxiang Zhou}\textsuperscript{\normalfont 2,*} \quad
  \textbf{Rong Xiong}\textsuperscript{\normalfont 1,2,*} \\[2pt]
  \textsuperscript{1}Zhejiang University \\
  \textsuperscript{2}Zhejiang Humanoid Robot Innovation Center \\
  \smash[b]{\begin{tabular}[t]{@{}c@{}}
    \textsuperscript{*}Corresponding authors. \\[2pt]
    {\normalfont\small\url{https://tract-lab.github.io/FOCUS-web/}}
  \end{tabular}}
}

\hypersetup{
  pdftitle={Anchoring Attention in Clutter: Mask-Guided Latent Distillation for Vision-Language-Action Towards Dexterous Grasping},
  pdfauthor={Xuanyu Zhang, He Li, Xinxiang Lou, Lu Chen, Yuxiang Cui, Zhongxiang Zhou, Rong Xiong},
  pdfkeywords={Dexterous Grasping, Vision-Language-Action, Visual Grounding}
}

\begin{document}
\maketitle

\begin{abstract}
Dexterous grasping in densely packed same-category clutter remains challenging for Vision-Language-Action (VLA) policies, in part because text-prompted VLAs provide little guidance for selecting among visually identical instances. Although external object masks are commonly provided as visual prompts, making the mask visible does not ensure that the policy maintains focused attention on the masked object---a failure mode especially consequential for dexterous-hand grasping, where small object-grounding errors directly affect contact and closure. We introduce \textbf{FOCUS}, a novel method for \textbf{F}oreground-aware, \textbf{O}bject-anchored \textbf{C}ontinuous grasping that \textbf{U}ses \textbf{S}elf-distillation to turn the mask from a passive visual prompt into a training-time alignment signal internalized by the action pathway. During training, FOCUS injects a mask-derived additive bias into the action expert's cross-attention and distills the resulting object-anchored representation into a bias-free pathway, encouraging control to be related to the mask-specified object region rather than object-specific appearance or illumination. To support sequential grasping across multiple objects, a lightweight progress query token co-attends with state and action tokens to trigger subtask termination and provide phase-aware context. In real-robot experiments at two levels of inter-instance distinguishability, FOCUS achieves 67.5\% average grasp success, surpassing all four VLA baselines by at least 26.9 percentage points, with the largest margin (+34.6 points) on the most challenging setting. Additional experiments confirm FOCUS's robustness to unseen object categories (58.4\%) and unseen lighting conditions (60.1\%), where all four baselines fall below 31\%.
\end{abstract}


\afterpage{\begin{figure*}[t]
\centering
\includegraphics[width=0.95\linewidth]{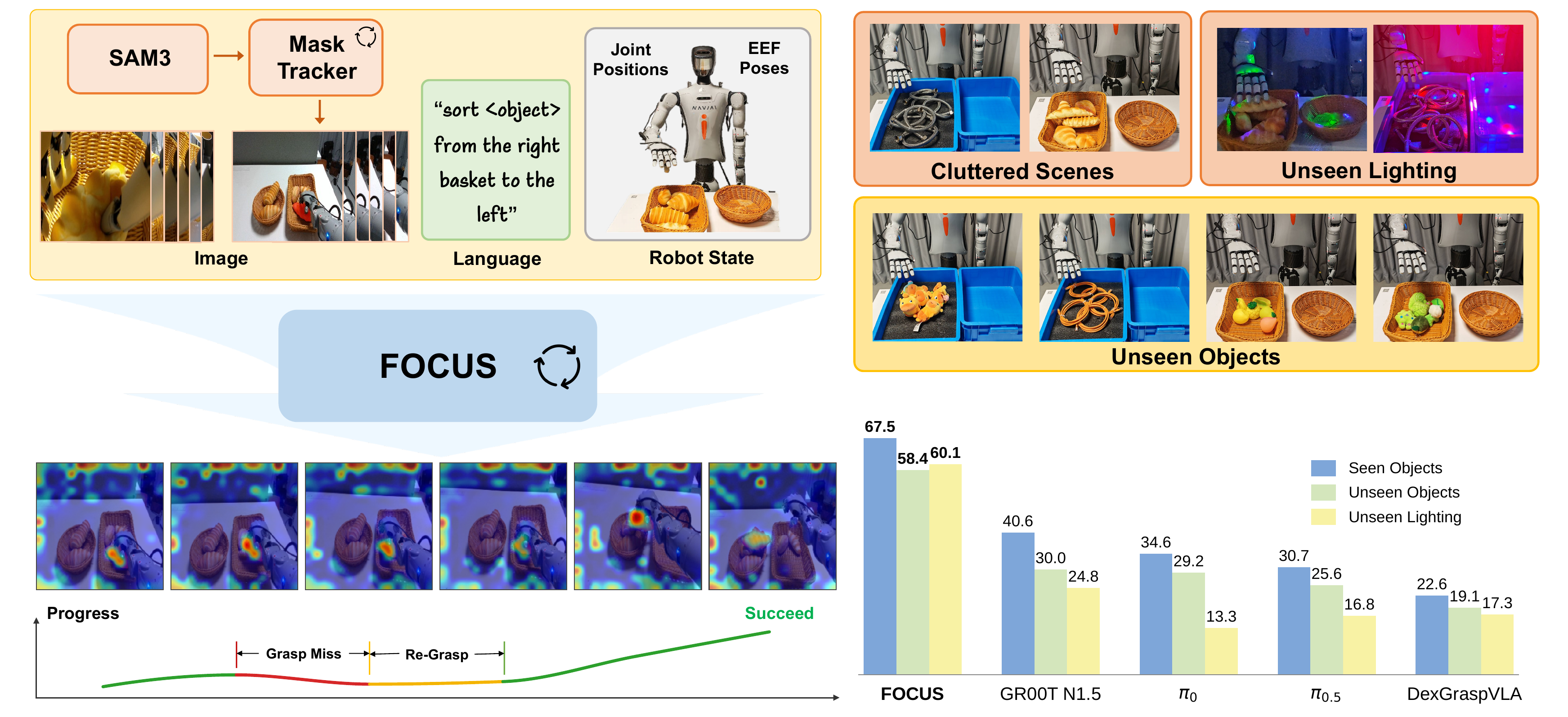}
\caption{FOCUS overview and real-robot evaluation. Taking a chest-camera view with a SAM3-tracked mask overlay, a wrist-camera image, a language instruction, and robot state as input, FOCUS policy drives closed-loop sequential sorting: mask-guided attention stays concentrated on the current object proposal, while the progress prediction drops on missed grasps and rises through re-grasping to success, signaling termination after the transfer. FOCUS exceeds VLA baselines across multiple evaluation settings.}
\label{fig:overview}
\end{figure*}}

\section{Introduction}
\label{sec:introduction}
Dexterous grasping in cluttered, densely packed scenes---as in industrial sorting, warehouse logistics, or household tidying---is an increasingly demanded real-world capability, yet remains a bottleneck for closed-loop manipulation systems~\citep{chen2025clutterdexgrasp,singh2024dextrahrgb,zhang2025robustdexgrasp}. Vision-Language-Action (VLA) models couple internet-scale vision-language priors with low-level action generation and have become a compelling foundation for such settings~\citep{brohan2022rt,brohan2023rt2,kim2024openvla,black2024pi_0,black2025pi}. However, in instance-ambiguous clutter, language alone provides little guidance for selecting and maintaining a stable grasp candidate. Practical systems therefore rely on an upstream perception module (e.g., open-vocabulary detection or promptable segmentation~\citep{liu2024groundingdino,kirillov2023sam,ravi2024sam2}) to propose an object region and provide it to the policy as a bounding box, point, or mask overlay (a colored region marking the target object on the image)~\citep{huang2025roboground,yu2025pointvla,zhong2026dexgraspvla,wang2026vpvla}.

Even with this mask cue, a deeper bottleneck remains: the cue is supplied at the input but does not directly constrain how the action expert allocates its internal attention. In dense same-category clutter, we observe \emph{weak object-region anchoring}: attention spreads across background patches and neighboring objects rather than concentrating on the object region (Section~\ref{sec:experiments}, Figure~\ref{fig:attention_mechanism}). The bottleneck is therefore not whether a mask is provided, but whether the \emph{action pathway internalizes} it. Moving from a single grasp to sequential sorting adds a second challenge: instances must be retrieved one at a time, so the system needs a subtask termination signal that drives upstream re-segmentation and downstream policy retargeting---a temporal signal that spatial anchoring alone cannot provide.

We propose FOCUS, a novel method that converts the mask from a passive prompt into a training-time alignment signal and pairs it with a progress query token for sequential subtask coordination (Figure~\ref{fig:overview}). FOCUS has three components. First, to address the anchoring bottleneck, a mask-derived additive bias modulates the action expert's cross-attention toward the masked object region during training. Second, to retain this anchoring at inference, latent self-distillation~\citep{baevski2022data2vec} transfers the resulting object-anchored representation into a parameter-shared, bias-free pathway. Third, addressing the sequential coordination need, a lightweight progress query token co-attends with state and action tokens to trigger subtask termination; sharing the action expert's object-anchored representation, the progress prediction itself benefits from the internalized mask cue.

We validate FOCUS on a real dexterous mobile manipulator in sequential sorting tasks under two levels of inter-instance distinguishability, including a Difficult level in which object instances are visually and geometrically identical. FOCUS attains 67.5\% average grasp success against 40.6\% / 34.6\% / 30.7\% / 22.6\% for four VLA baselines, including a mask-conditioned baseline, with a 34.6-point Difficult-level margin; the advantage is preserved on held-out object categories and unseen lighting. Qualitative and quantitative attention analyses further show that FOCUS concentrates action attention more strongly on the masked object region than standard fine-tuning or mask-overlay-only adaptation.

Our main contributions are:
\begin{enumerate}[leftmargin=*, labelsep=0.6em, align=left, itemsep=0.15em, topsep=0.2em, parsep=0pt, partopsep=0pt]
    \item A mask-guided attention alignment mechanism that converts segmentation masks into explicit action-path supervision via cross-attention bias and biased-to-unbiased latent self-distillation, improving object-region anchoring without changing the deployment interface.
    \item A progress query token in the action expert's self-attention sequence, where it co-attends with state and action tokens to learn a representation that supports next-object proposal selection and provides phase-aware context for action generation.
    \item An extensive real-robot evaluation of sequential sorting across two levels of inter-instance distinguishability and two generalization settings. FOCUS consistently outperforms RGB-only and mask-conditioned VLA baselines, with qualitative and quantitative attention analyses supporting stronger object-region grounding.
\end{enumerate}

\section{Related Work}
\label{sec:related_work}

\paragraph{Dexterous Grasping in Cluttered Scenes.}
Classical grasp pipelines detect candidate poses via geometric analysis, analytic quality metrics, or learned scoring on point clouds~\citep{tenpas2017gpd,mahler2017dexnet2,fang2020graspnet,sundermeyer2021contact,fang2023anygrasp} and execute them open-loop; for dexterous hands, analytic synthesis and differentiable optimization can produce diverse hand poses but typically assume isolated objects or light clutter~\citep{li2023gendexgrasp,wan2023unidexgrasppp,liu2021forceclosure,turpin2022graspd,turpin2023fastgraspd,xu2024dipgrasp}. Recent end-to-end learning approaches---reinforcement learning in simulation~\citep{singh2024dextrahrgb,chen2025clutterdexgrasp,zhang2025robustdexgrasp} and imitation from teleoperation~\citep{arunachalam2023dime,zhao2023aloha,jiang2025dexmimicgen}---improve robustness in physically dense scenes, yet do not resolve the instance-selection challenge among visually identical objects. The policy must maintain focus on one graspable object region through approach, contact, and closure, while avoiding background contact and multi-object retrieval. We address this setting by anchoring the action pathway of a pretrained VLA to a mask-specified object region.

\paragraph{Visual Grounding in Vision-Language-Action (VLA) Policies.}
VLA policies inherit broad vision-language priors for robotic control. Early systems such as RT-1~\citep{brohan2022rt}, RT-2~\citep{brohan2023rt2}, and OpenVLA~\citep{kim2024openvla} tokenize discretized actions, while continuous-action successors~\citep{team2024octo,black2024pi_0,black2025pi} use diffusion- or flow-matching-based action heads~\citep{chi2024diffusionpolicy,lipman2023flow} for smoother chunk generation. However, in dense same-category clutter, semantic priors correctly identify the object category but provide no guidance for selecting and persistently tracking a single instance among visually identical peers. This motivates grounding interfaces that feed object-region cues to VLA policies through auxiliary mask inputs~\citep{huang2025roboground,zhong2026dexgraspvla} and rendered RGB visual prompts~\citep{yu2025pointvla,wang2026vpvla}. Yet such input-level augmentations provide grounding as an inference-time conditioning signal, rather than internalizing the mask-guided grounding signal within the action pathway itself. Other approaches insert intermediate reasoning steps before action decoding---embodied chain-of-thought~\citep{zawalski2024robotic}, sub-goal image prediction~\citep{zhao2025cot}, and autoregressive box and grasp-pose tokens~\citep{deng2025graspvla}---but introduce additional inference-time computation. FOCUS targets this policy-side bottleneck: mask-guided attention routing is distilled during training into the bias-free action pathway, shaping the visuomotor representation used for control without additional inference-time reasoning.

\paragraph{Progress-Aware Control and Sub-task Transitions.}
Learned transition signals have been incorporated into robot policies at different levels. RT-2 includes termination in its discretized action vocabulary~\citep{brohan2023rt2}, TaKSIE uses recurrent visual-progress features to trigger visual-subgoal generation~\citep{kang2025tasksie}, the subgoal-transition models of Jain et al.~\citep{jain2025subgoaltransitions} estimate when to advance between learned subgoals, and PALM jointly predicts continuous within-subtask progress and actions from affordance-conditioned representations~\citep{liu2026palm}. FOCUS addresses a narrower role in sequential grasping: a single query in the action expert estimates observation-conditioned within-subtask progress, supplies phase context to action tokens, and triggers subtask transitions, without introducing a separate subgoal generator or requiring semantic-stage annotations.

\section{Method}
\label{sec:method}
\begin{figure*}[t]
\centering
\includegraphics[width=\linewidth]{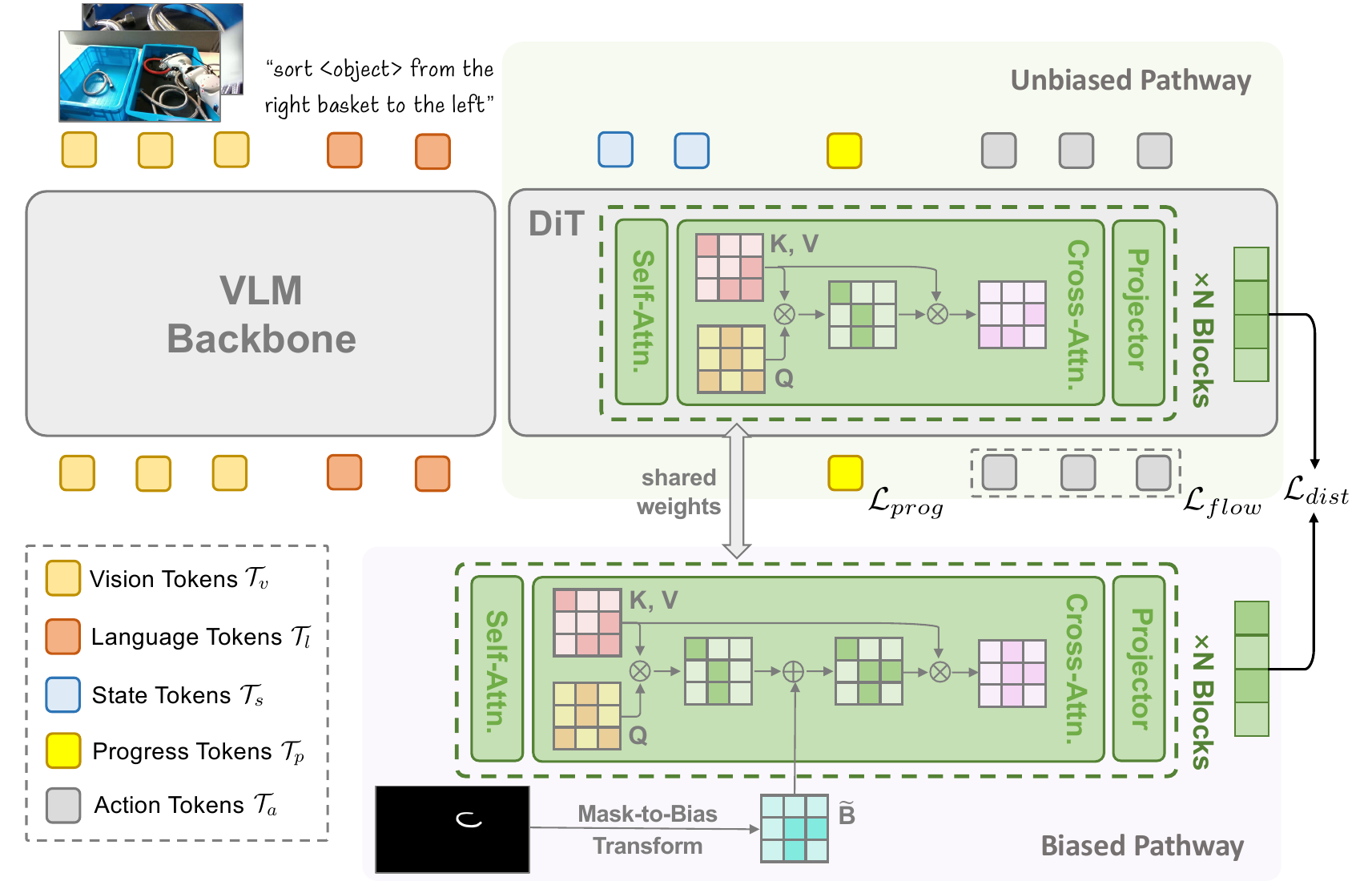}
\caption{Two parameter-shared pathways of the action expert (flow-matching DiT): a \emph{biased} pathway with the mask-derived attention bias $\tilde{\mathbf{B}}$, and an \emph{unbiased} pathway matching deployment. Under stop-gradient on the biased pathway, $\mathcal{L}_{\text{dist}}$ aligns the unbiased pathway's hidden states with the biased one's; a progress query token $\mathcal{T}_p$ co-attends with state and action tokens, supervised by $\mathcal{L}_{\text{prog}}$. The total training objective is $\mathcal{L}_{\text{flow}} + \lambda_{\text{dist}}\mathcal{L}_{\text{dist}} + \lambda_{\text{prog}}\mathcal{L}_{\text{prog}}$. At inference, only the unbiased pathway is used.}
\label{fig:framework}
\end{figure*}

FOCUS uses mask-guided attention routing during training, then distills the induced object-region alignment into the unbiased deployment pathway. Concretely, in the action expert of GR00T N1.5 (3B)~\citep{gr00tn15_2025,gr00tn1_2025}---a flow-matching DiT~\citep{lipman2023flow} where action tokens $\mathcal{T}_a$ and proprioceptive state tokens $\mathcal{T}_s$ cross-attend to a vision-language token sequence $\mathcal{T}_{vl}$---FOCUS adds (i) a mask-guided soft attention prior on the action expert's cross-attention (Section~\ref{sec:mask_prior}), (ii) a biased-to-unbiased latent self-distillation that preserves the object-region alignment after the bias is removed at inference (Section~\ref{sec:distill}), and (iii) a progress query token that supplies subtask termination and phase-aware context (Section~\ref{sec:progress}). Figure~\ref{fig:framework} illustrates the overall architecture.

\paragraph{Mask-guided soft attention prior.}
\label{sec:mask_prior}
Given an object mask from an upstream perception module, we convert it into a spatial modulation prior on the action expert's cross-attention. Let the camera-view visual tokens in $\mathcal{T}_{vl}$ correspond to $N$ image patches; we downsample the mask to a weight vector $\mathbf{g} \in [0, 1]^N$ aligned with these patches and add a per-patch bias $\tilde{\mathbf{B}} \in \mathbb{R}^N$ to the action-to-vision attention logits:
$$B_i = \alpha\log\!\big(g_i(1-\eta)+\eta\big),\; \tilde{B}_i = B_i - \tfrac{1}{N}\textstyle\sum_{j} B_j,\; \text{Attention}(Q, K, V) = \mathrm{softmax}\!\left(\tfrac{QK^\top}{\sqrt{d}} + \tilde{\mathbf{B}}\right)V,$$
where $\eta \in (0,1)$ is a per-patch floor that keeps the bias finite and retains background context, and $\alpha$ follows a three-stage cosine schedule (Appendix~\ref{app:training_details}). The logarithmic form makes $\tilde{\mathbf{B}}$ equivalent, after softmax, to multiplicative reweighting in the attention-probability domain, suppressing irrelevant background patches while preserving local context around the object proposal; the zero-mean normalization keeps the bias centered across samples. In parallel, we render the mask as a semi-transparent object overlay $I_o$ on the chest-camera image, with per-trajectory color randomization at training and contrast-adaptive tint selection at deployment (Appendix~\ref{app:overlay_rendering}).

\paragraph{Latent self-distillation via biased-to-unbiased consistency.}
\label{sec:distill}
To internalize the alignment induced by $\tilde{\mathbf{B}}$, we run two parameter-shared forward pathways: a \emph{biased} pathway with $\tilde{\mathbf{B}}\neq0$ that yields an explicitly object-aligned representation, and an \emph{unbiased} pathway with $\tilde{\mathbf{B}}=0$ that matches the deployment configuration. The biased pathway is computed under stop-gradient, providing a fixed teacher representation while gradients flow only through the unbiased pathway. This asymmetry pulls the deployment-compatible representation toward the object-aligned one and avoids collapsing both pathways to an uninformative consensus. We apply the distillation objective at the final-layer hidden states of the action tokens $\mathcal{T}_a$ and the progress token $\mathcal{T}_p$ (introduced in Section~\ref{sec:progress}), using the mean token-wise cosine distance:
$$\mathcal{L}_{\text{dist}} = 1 - \frac{1}{|\mathcal{T}_a|+|\mathcal{T}_p|}\sum_{i=1}^{|\mathcal{T}_a|+|\mathcal{T}_p|} \frac{\mathbf{h}_i^{u} \cdot \mathbf{h}_i^{b}}{\|\mathbf{h}_i^{u}\|_2 \, \|\mathbf{h}_i^{b}\|_2}.$$
At deployment the explicit bias is removed, but the unbiased pathway retains a latent space that preserves spatially grounded action features without any additional inference-time routing.

\paragraph{Progress query token.}
\label{sec:progress}
To support subtask transitions in multi-step grasping, we append a learnable progress query token $\mathcal{T}_p$ to the action expert's self-attention sequence, where it co-attends with $\mathcal{T}_s$ and $\mathcal{T}_a$. This bidirectional connectivity lets $\mathcal{T}_p$ aggregate phase context from the state--action manifold and (via cross-attention) from $\mathcal{T}_{vl}$, while feeding progress-aware features back into the action tokens. Its hidden state is decoded by a small MLP into a scalar progress estimate $p_t\in[0,1]$. Because the training trajectories are successful expert demonstrations, normalized time provides a dense proxy for within-subtask progress; we supervise $p_t$ against $y_t=t/T$ with $\mathcal{L}_{\text{prog}} = \| p_t - y_t \|^2$. Its primary role is closed-loop termination; its secondary effect on action generation is quantified in Section~\ref{sec:experiments}, and its behavior under pauses and re-grasping is evaluated in Appendix~\ref{app:progress_events}.

We optimize FOCUS end-to-end with $\mathcal{L}_{\text{total}} = \mathcal{L}_{\text{flow}} + \lambda_{\text{dist}}\mathcal{L}_{\text{dist}} + \lambda_{\text{prog}}\mathcal{L}_{\text{prog}}$. The bias-strength schedule $\alpha(t)$ (Appendix~\ref{app:training_details}) lets the unbiased pathway transition smoothly from bias-guided alignment to autonomous object-grounded control from the overlay alone.

\subsection{Inference and Closed-Loop Execution}
At each control cycle the policy receives a chest-camera view with a SAM3-generated~\citep{carion2025sam3segmentconcepts} mask overlay $I_o$ for the current object proposal (refreshed frame-by-frame with a per-subtask tint chosen for contrast; details in Appendices~\ref{app:overlay_rendering} and~\ref{app:deployment}) and a wrist-camera image, and emits an action chunk together with a scalar progress estimate $p_t$. Once $p_t>\tau_p$ with $\tau_p=0.8$, the grasp-and-transfer subtask is deemed complete and SAM3 proposes the next object. Controlled policy latency is reported in Appendix~\ref{app:efficiency}.

\section{Experiments}
\label{sec:experiments}
\subsection{Experimental Setup}
\textbf{Hardware.} Our platform is a dual-arm mobile manipulator with dexterous hands; we control only the right side---a 7-DoF arm and a 6-DoF dexterous hand together with 2 waist joints (pitch/yaw), giving a 15-D continuous action space. Visual observations come from a chest camera (global view) and a wrist camera (hand-centric feedback).

\begin{figure*}[!t]
\centering
\begin{subfigure}[t]{0.32\linewidth}
  \centering
  \includegraphics[width=\linewidth]{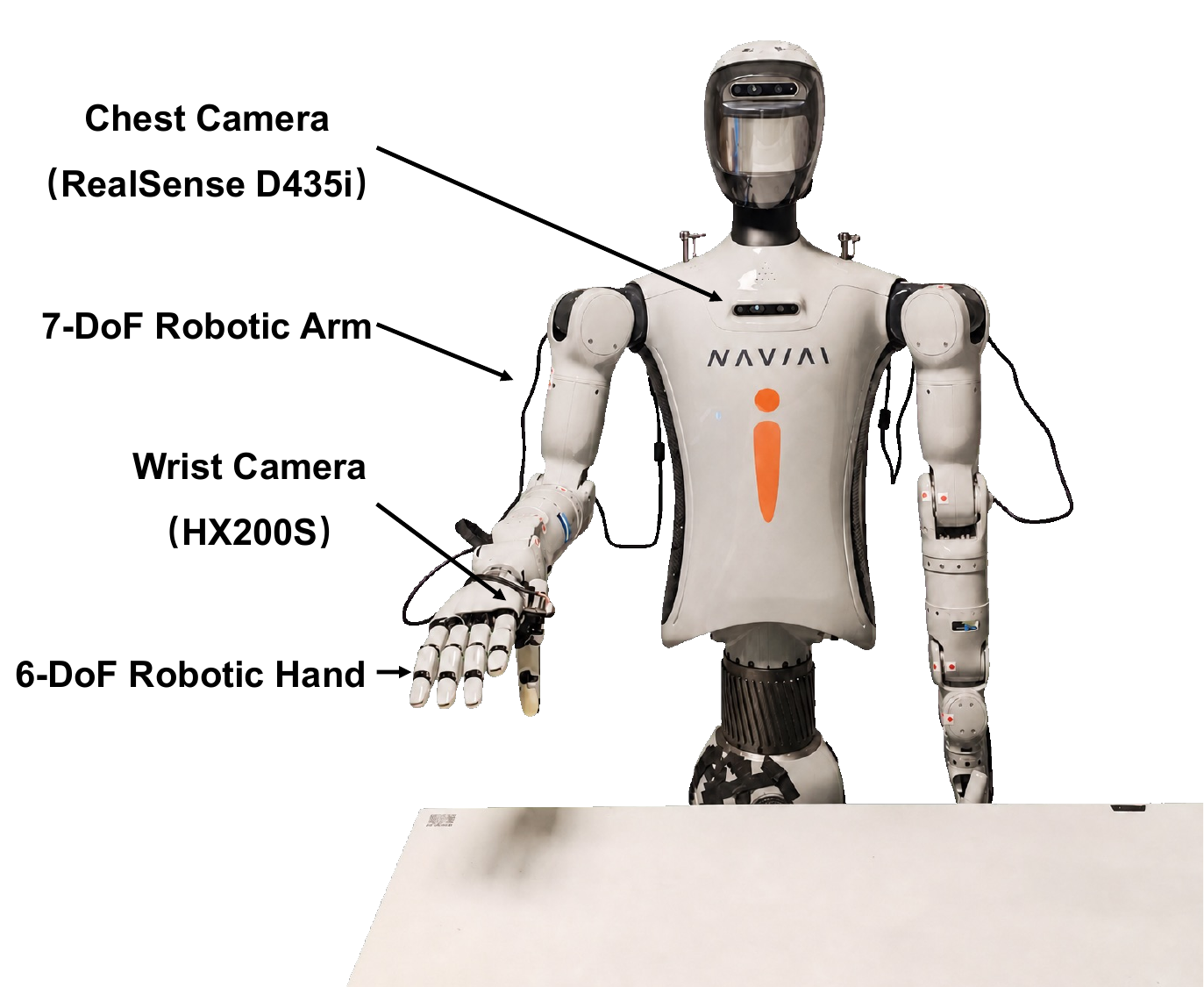}
  \caption{Hardware platform}
  \label{fig:hardware_platform}
\end{subfigure}
\hfill
\begin{subfigure}[t]{0.66\linewidth}
  \centering
  \includegraphics[width=\linewidth]{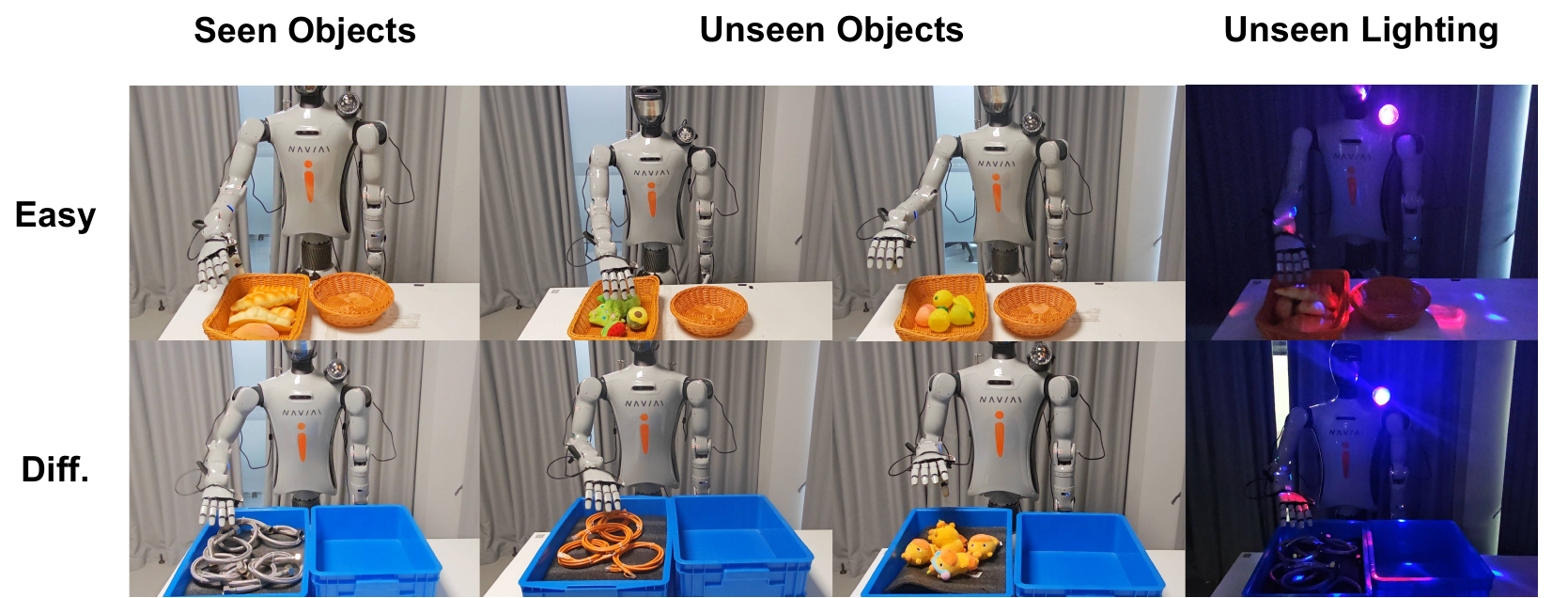}
  \caption{Task design and object categories}
  \label{fig:task_design}
\end{subfigure}
\caption{Experimental setup. (a) Real-robot platform. (b) Sorting task with \textbf{Seen}/\textbf{Unseen} objects and \textbf{Unseen lighting} (held out for generalization). Easy instances vary in shape, color, or texture; Difficult instances are visually and geometrically identical.}
\label{fig:objects}
\end{figure*}

\textbf{Task protocol and difficulty levels.} Each episode contains $N$ same-category object instances, with $N$ sampled uniformly from $\{5,6,7\}$ per episode and the $N$ instances randomly piled into a large bin; the robot must then grasp them one by one and transfer them into a packing box until the bin is cleared or $4N$ attempts have been made. Success therefore depends on whether the policy can maintain object-level focus through approach and closure, while avoiding background contact and multi-object retrieval. To study how inter-instance distinguishability affects single-object grasp consistency, we define two difficulty levels by inter-instance distinguishability (Figure~\ref{fig:objects}):
\begin{itemize}[leftmargin=*, labelsep=0.6em, itemsep=1pt, topsep=1pt]
    \item \textbf{Easy.} Instances share the category label but differ substantially in shape, color, or texture, and may provide additional cues for object selection.
    \item \textbf{Difficult.} Instances are visually and geometrically identical; appearance provides minimal cues for separating adjacent instances.
\end{itemize}
Each method-difficulty combination within a given setting is evaluated over 10 episodes with independently randomized piles; for the unseen-objects setting these are evenly split across two held-out object categories (5 episodes per category). Episode counts and total grasp attempts for each condition (varying with method due to different attempt budgets per episode) are listed in Appendix~\ref{app:stats}.

\begin{figure*}[!t]
\centering
\begin{subfigure}[t]{0.49\linewidth}
  \centering
  \includegraphics[height=0.13\textheight]{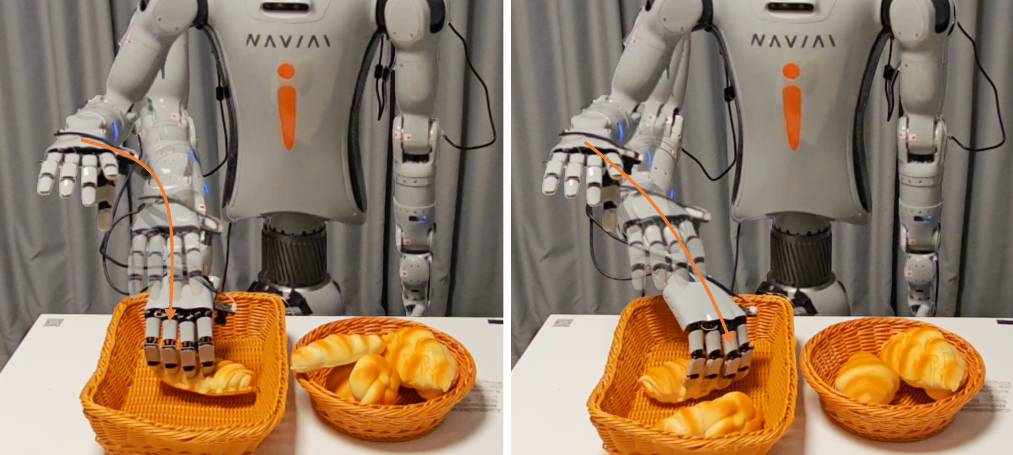}
  \caption{Efficient grasping under pose variation}
  \label{fig:efficient_grasping}
\end{subfigure}
\hfill
\begin{subfigure}[t]{0.49\linewidth}
  \centering
  \includegraphics[height=0.13\textheight]{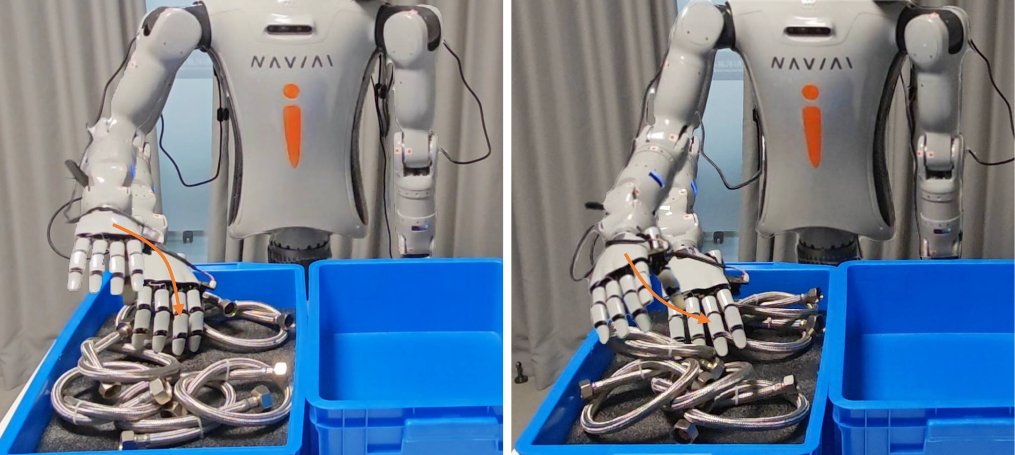}
  \caption{Clutter-aware grasping under partial occlusion}
  \label{fig:clutter_aware_grasping}
\end{subfigure}
\caption{Qualitative FOCUS executions. (a) The policy adjusts approach angle to an elongated bread and grasps from an easier side. (b) For a partially occluded object in stacked clutter, it approaches from the exposed side, avoiding neighbor contact.}
\label{fig:qualitative_grasping}
\end{figure*}

\textbf{Data collection.} We collect $\sim$3.5 hours of teleoperated demonstrations across 1{,}800 trajectories (900 per difficulty level) with randomized object poses, clutter layouts, and container poses. Each trajectory records the chest-camera image (with the corresponding binary mask and its rendered overlay), the wrist-camera image, proprioceptive state, action, and a normalized progress label $y_t=t/T$; training-time masks are initialized with manual point prompts to SAM3, then propagated with SAM3 tracking (Appendix~\ref{app:overlay_rendering}). All methods are fine-tuned on the same demonstration set.

\textbf{Baselines and input information.} All methods receive chest and wrist RGB views, robot state, and the same language instruction. Mask-conditioned methods additionally use an external mask extractor. Following its official implementation, DexGraspVLA receives the extracted mask as a separate input channel, whereas FOCUS configurations render the SAM3 mask on the chest RGB. We compare against four representative VLAs: (1) \textbf{GR00T N1.5}~\citep{gr00tn15_2025}, the same pretrained backbone with standard fine-tuning; (2) \textbf{$\pi_0$}~\citep{black2024pi_0}, a flow-matching continuous-action VLA; (3) \textbf{$\pi_{0.5}$}~\citep{black2025pi}, its open-world-generalization extension; and (4) \textbf{DexGraspVLA}~\citep{zhong2026dexgraspvla}, a mask-conditioned dexterous-grasping VLA. All baselines are evaluated under the same closed-loop sequential-sorting protocol as FOCUS.

\textbf{Evaluation.} We report the \textbf{mean per-episode success rate}: the fraction of attempts in each episode that grasp and transfer exactly one object under a strict single-object criterion (missed grasps and multi-object retrievals are both counted as failures), averaged across episodes (and across the two held-out object categories for the unseen-objects setting). This per-episode aggregation prevents high-attempt episodes from dominating shorter ones, and reflects single-object grasp consistency by penalizing both unfocused approaches and unintended neighbor contact. For grasp-success results, subscripts indicate the standard error $\mathrm{SE}=\sigma_{\text{ep}}/\sqrt{n_{\text{ep}}}$ with $\sigma_{\text{ep}}$ the per-episode sample std (ddof$=$1); the Avg.\ column propagates error as $\mathrm{SE}_{\text{Avg}}=\tfrac{1}{2}\sqrt{\mathrm{SE}_{\text{Easy}}^2+\mathrm{SE}_{\text{Diff}}^2}$.

\Needspace{5\baselineskip}
\subsection{Main Results on Cluttered Multi-Instance Sorting}
\begin{wraptable}{r}{0.48\linewidth}
\centering
\caption{In-domain per-episode success rate (mean $\pm$ SE, \%; App.~\ref{app:stats}).}
\label{tab:main_results}
\renewcommand{\arraystretch}{1.05}
\resizebox{\linewidth}{!}{\begin{tabular}{@{}lccc@{}}
\toprule
Method & Easy & Diff. & Avg. \\
\midrule
GR00T N1.5 & 48.87$_{\pm 3.00}$ & 32.27$_{\pm 1.87}$ & 40.57$_{\pm 1.77}$ \\
$\pi_0$ & 37.73$_{\pm 3.93}$ & 31.40$_{\pm 1.93}$ & 34.56$_{\pm 2.19}$ \\
$\pi_{0.5}$ & 34.49$_{\pm 1.71}$ & 26.93$_{\pm 2.90}$ & 30.71$_{\pm 1.68}$ \\
DexGraspVLA & 24.27$_{\pm 1.72}$ & 20.99$_{\pm 1.20}$ & 22.63$_{\pm 1.05}$ \\
\textbf{FOCUS} & \textbf{68.05}$_{\pm 3.97}$ & \textbf{66.91}$_{\pm 2.62}$ & \textbf{67.48}$_{\pm 2.38}$ \\
\bottomrule
\end{tabular}}
\end{wraptable}
Table~\ref{tab:main_results} reports in-domain success. FOCUS reaches 67.5\% average success, improving over GR00T N1.5 (40.6\%), $\pi_0$ (34.6\%), $\pi_{0.5}$ (30.7\%), and the mask-conditioned DexGraspVLA (22.6\%) by 26.9--44.9 points. The margin is largest at the Difficult level (+34.6 over GR00T N1.5), where object instances are visually identical and tightly entangled. DexGraspVLA underperforms GR00T N1.5 despite explicitly consuming a mask as an input channel; together with the mask-overlay configuration in Table~\ref{tab:ablation}, this suggests that input-level mask access is insufficient for object-region anchoring in our dexterous, contact-rich setting.

Qualitative executions (Figure~\ref{fig:qualitative_grasping}) illustrate how these gains appear at control time. In the elongated-bread example, FOCUS adapts the arm approach angle to the object's pose and grasps from an easier side rather than approaching frontally; in the stacked-clutter example, the policy approaches a partially occluded object proposal from its exposed side, reducing incidental contact with neighbors. Object-region anchoring thus manifests not only as localization but also as pose-aware and clutter-aware grasp execution.

FOCUS changes by only 1.1 points from Easy to Difficult, whereas the RGB-only baselines degrade more substantially; DexGraspVLA remains low in both settings. This pattern is consistent with stronger use of the mask-specified object region rather than residual object appearance.

\begin{table}[!ht]
\centering
\caption{Zero-shot generalization to unseen object categories and unseen lighting (mean $\pm$ SE, \%; App.~\ref{app:stats}).}
\label{tab:generalization}
\renewcommand{\arraystretch}{1.05}
\resizebox{\linewidth}{!}{\begin{tabular}{@{}l ccc ccc@{}}
\toprule
\multirow{2}{*}{Method} & \multicolumn{3}{c}{Unseen Objects} & \multicolumn{3}{c}{Unseen Lighting} \\
\cmidrule(lr){2-4}\cmidrule(lr){5-7}
& Easy & Diff. & Avg. & Easy & Diff. & Avg. \\
\midrule
GR00T N1.5 & 37.13$_{\pm 5.10}$ & 22.95$_{\pm 2.31}$ & 30.04$_{\pm 2.80}$ & 27.27$_{\pm 1.23}$ & 22.27$_{\pm 3.01}$ & 24.77$_{\pm 1.62}$ \\
$\pi_0$ & 31.26$_{\pm 3.37}$ & 27.07$_{\pm 3.18}$ & 29.16$_{\pm 2.32}$ & 15.00$_{\pm 1.11}$ & 11.67$_{\pm 1.84}$ & 13.33$_{\pm 1.08}$ \\
$\pi_{0.5}$ & 28.95$_{\pm 2.52}$ & 22.27$_{\pm 2.25}$ & 25.61$_{\pm 1.69}$ & 15.83$_{\pm 1.04}$ & 17.78$_{\pm 1.15}$ & 16.81$_{\pm 0.77}$ \\
DexGraspVLA & 20.65$_{\pm 2.49}$ & 17.62$_{\pm 1.37}$ & 19.13$_{\pm 1.42}$ & 19.56$_{\pm 2.68}$ & 15.00$_{\pm 1.67}$ & 17.28$_{\pm 1.58}$ \\
\textbf{FOCUS} & \textbf{60.48}$_{\pm 3.58}$ & \textbf{56.39}$_{\pm 3.89}$ & \textbf{58.43}$_{\pm 2.64}$ & \textbf{57.98}$_{\pm 2.47}$ & \textbf{62.29}$_{\pm 3.60}$ & \textbf{60.14}$_{\pm 2.18}$ \\
\bottomrule
\end{tabular}}
\par\vspace{\floatsep}
\caption{Cumulative training-component ablation on GR00T N1.5 (mean per-episode success rate, \% $\pm$ SE). $^\dagger$The attention bias is retained at inference only in this configuration; distilled configurations use the standard inference pathway.}
\label{tab:ablation}
\small
\renewcommand{\arraystretch}{1.05}
\setlength{\tabcolsep}{2.5pt}
\begin{tabular}{@{}lccc@{}}
\toprule
Configuration & Easy & Diff. & Avg. \\
\midrule
GR00T N1.5 & 48.87$_{\pm 3.00}$ & 32.27$_{\pm 1.87}$ & 40.57$_{\pm 1.77}$ \\
$+$ Mask overlay & 51.04$_{\pm 3.01}$ & 38.05$_{\pm 2.92}$ & 44.55$_{\pm 2.10}$ \\
$+$ Attention bias$^\dagger$ & 66.15$_{\pm 3.85}$ & 62.45$_{\pm 2.59}$ & 64.30$_{\pm 2.30}$ \\
$+$ Latent distillation & 64.43$_{\pm 4.29}$ & 59.12$_{\pm 4.09}$ & 61.77$_{\pm 2.96}$ \\
$+$ Progress conditioning (FOCUS) & \textbf{68.05}$_{\pm 3.97}$ & \textbf{66.91}$_{\pm 2.62}$ & \textbf{67.48}$_{\pm 2.38}$ \\
\bottomrule
\end{tabular}
\end{table}
\begin{figure*}[!htbp]
  \centering
  \includegraphics[width=0.98\linewidth]{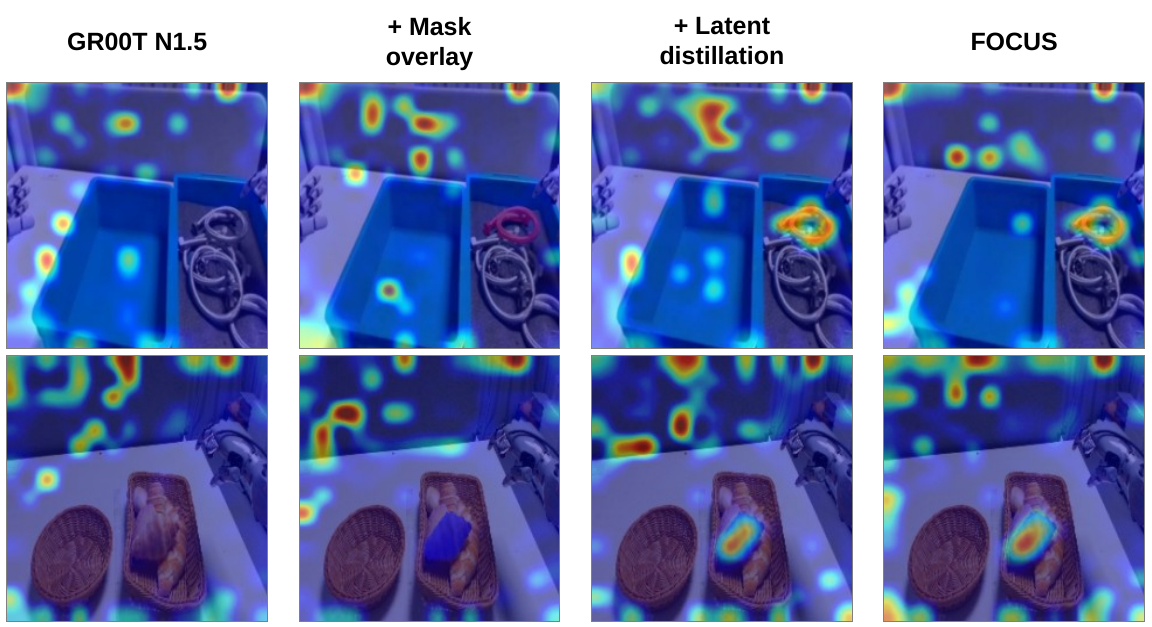}
  \caption{First-step action-attention heatmaps on the chest-camera observation. Standard fine-tuning and mask-overlay-only adaptation show no clear attention on the object region; FOCUS concentrates attention on the masked object while preserving attention to surrounding context for collision avoidance and placement. The latent-distillation configuration and FOCUS yield visually similar maps; progress conditioning primarily affects action quality rather than first-step attention (Table~\ref{tab:ablation}).}
  \label{fig:attention_mechanism}
\end{figure*}

\begin{table}[!htbp]
\centering
\caption{Target-to-background attention ratio (TBR$\uparrow$) on 100 randomly sampled test observations per difficulty split. Subscripts denote 95\% bootstrap confidence intervals.}
\label{tab:tbr}
\small
\renewcommand{\arraystretch}{1.05}
\setlength{\tabcolsep}{7pt}
\begin{tabular}{@{}lcc@{}}
\toprule
Configuration & Easy & Diff. \\
\midrule
GR00T N1.5 & $0.55_{[0.52,0.58]}$ & $0.53_{[0.49,0.56]}$ \\
$+$ Mask overlay & $0.57_{[0.52,0.63]}$ & $0.41_{[0.38,0.45]}$ \\
\textbf{FOCUS} & $\mathbf{2.17}_{[2.08,2.26]}$ & $\mathbf{3.22}_{[3.05,3.41]}$ \\
\bottomrule
\end{tabular}
\end{table}

\subsection{Generalization to Unseen Objects and Unseen Lighting}
Table~\ref{tab:generalization} tests two zero-shot settings (Figure~\ref{fig:objects}): held-out object categories that preserve the difficulty level's visual distinguishability but differ in shape, material, and appearance; and held-out illumination on the training objects.

\textbf{Unseen objects.} FOCUS attains 58.4\% average success vs.\ 30.0\% / 29.2\% / 25.6\% / 19.1\% for GR00T N1.5 / $\pi_0$ / $\pi_{0.5}$ / DexGraspVLA. The Difficult-level advantage persists (56.4\% vs.\ 27.1\% for the best baseline), indicating that the learned object-region anchoring generalizes to new instances rather than fitting to specific training objects.

\textbf{Unseen lighting.} FOCUS reaches 60.1\% average success while RGB-only baselines fall to 13.3--24.8\% and DexGraspVLA to 17.3\%, a substantially larger drop than for FOCUS. The persistent gap from DexGraspVLA is consistent with the interpretation that input-level mask conditioning alone may not produce an object-aligned visuomotor representation that transfers across illumination.

\subsection{Ablations and Object-Region Grounding Analysis}
Table~\ref{tab:ablation} incrementally isolates the mask overlay, attention routing, latent distillation, and progress conditioning. In the latent-distillation configuration, the progress token still predicts termination, but action and state tokens cannot attend to it; the final row enables this progress-to-action connection. The mask overlay raises average success from 40.57\% to 44.55\%, while direct attention routing reaches 64.30\% (+19.75 points over the overlay). Latent distillation retains most of this gain without explicit routing at inference (61.77\%), and progress conditioning further raises performance to 67.48\%. Controlled profiling shows that distillation increases training time by 1.87\%, whereas retaining direct attention routing increases policy inference latency by 45.89\% (Appendix~\ref{app:efficiency}).

To quantify object-region grounding, Table~\ref{tab:tbr} reports the ratio of geometric-mean action-to-vision cross-attention over target and background regions (details in Appendix~\ref{app:tbr_details}). FOCUS reaches 2.17/3.22 on Easy/Difficult, whereas standard fine-tuning and mask-overlay-only adaptation remain below one. The quantitative ratios and qualitative heatmaps in Figure~\ref{fig:attention_mechanism} provide complementary evidence: input-level masking alone does not consistently concentrate action attention on the selected object, while FOCUS produces substantially stronger object-region anchoring.

\FloatBarrier

\section{Conclusion}
\label{sec:conclusion}
We presented FOCUS, a method that turns object masks into training-time alignment signals for dexterous VLA policies. It combines mask-guided cross-attention bias with latent self-distillation to align the action expert with the selected object region. A progress query token provides phase-aware context for action generation and supports closed-loop subtask transitions. The resulting policy improves object-region anchoring inside the action expert, as supported by real-robot performance and quantitative attention analysis. In real-robot experiments on a dexterous mobile manipulator, FOCUS outperforms RGB-only VLAs and a mask-conditioned baseline on in-domain, unseen-object, and unseen-lighting settings, with the largest gains where instance ambiguity is most severe. These results suggest a broader design principle for closed-loop manipulation: task-relevant cues identified upstream should be internalized by the action pathway, not only exposed at the input.

\section{Limitations}
\label{sec:limitations}
FOCUS remains dependent on the quality of the upstream object mask. Appendix~\ref{app:mask_perturbations} shows robustness to controlled erosion and truncation, while more severe mask corruption remains outside the present evaluation. In addition, the failure analysis in Appendix~\ref{app:failure_modes} suggests that, in our evaluated setting, much of the remaining headroom lies in contact-level dexterous execution rather than coarse object-region grounding. In particular, slippage during lift, boundary-region grasps between touching instances, and height misalignment point to limitations that are not fully addressed by mask-level object-region anchoring alone. Potential improvements include tactile feedback for closed-loop grasp adjustment, grasp-point-level affordance priors, and depth-sensitive observations. The current demonstrations and evaluation cover a limited range of dexterous grasp behaviors and focus on sequential sorting, leaving richer hand behaviors such as adaptive finger gaiting, in-hand reorientation, functional grasping, and non-prehensile manipulation to future work. Finally, although the mask-guided bias and latent distillation may extend to VLA architectures with attention-based action decoders, empirical verification across multiple backbones remains future work.

\acknowledgments{This work was supported by the Joint Funds of the National Natural Science Foundation of China (Grant No.~U25A6013).}
\bibliography{references}

\clearpage
\appendix

\section{Implementation Details}
\label{app:impl_details}

\subsection{Model Architecture and Training}
\label{app:training_details}

\paragraph{Architecture and optimization.} We fine-tune NVIDIA GR00T N1.5 (3B)~\citep{gr00tn15_2025,gr00tn1_2025}, which couples an Eagle-2.5 vision-language backbone with a flow-matching DiT action head. We freeze the language-model backbone and train the vision tower, multimodal projector, and DiT action head. We optimize with AdamW ($(\beta_1,\beta_2){=}(0.95, 0.999)$, weight decay $1{\times}10^{-5}$, base learning rate $1{\times}10^{-4}$, cosine decay with a 5\% linear warm-up) for $30{,}000$ steps on $2{\times}$ NVIDIA H800 GPUs at global batch size $64$.

\paragraph{Mask-guided bias schedule.} We adaptively pool the binary chest-camera mask to the visual-token grid and construct the zero-centered logit bias as defined in Section~\ref{sec:mask_prior}, using a floor of $\eta=0.05$. The bias strength follows a three-stage cosine schedule: $\alpha$ increases from $0$ to $2.0$ over the first $10{,}000$ steps, decreases to $0$ over steps $10{,}000$--$25{,}000$, and remains at $0$ for the final $5{,}000$ steps.

\paragraph{Training objectives.} We retain GR00T N1.5's flow-matching velocity loss, using $u=(s-\tau)/s$, $\tau\sim\mathrm{Beta}(1.5,1.0)$, and $s=0.999$. The flow loss is the masked MSE between the predicted velocity and $a-\epsilon$. Each training step uses two parameter-shared DiT passes: an unbiased student pass with gradients and a biased teacher pass under stop-gradient. Distillation minimizes the mean cosine distance between their final-layer action- and progress-token states. An embodiment-specific MLP decodes the progress token into $p_t\in[0,1]$, supervised by MSE against $t/T$. We use $\lambda_{\mathrm{dist}}=0.2$ and $\lambda_{\mathrm{prog}}=0.1$ in the objective defined in Section~\ref{sec:method}.

\subsection{Mask Generation and Overlay Rendering}
\label{app:overlay_rendering}

\textbf{Training masks.} For each teleoperated grasp-transfer trajectory, we initialize the target object in the first frame with a small number of manual point prompts---positive clicks on the target, plus negative clicks to exclude neighboring instances---and use point-prompted SAM3 to obtain the initial binary mask. The mask is then propagated to subsequent frames with the SAM3 video tracker, producing the per-frame masks used for both the mask-guided attention bias and the rendered overlay.

\begin{figure}[!htbp]
  \centering
  \begin{minipage}[t]{0.32\linewidth}
    \centering
    \includegraphics[width=\linewidth]{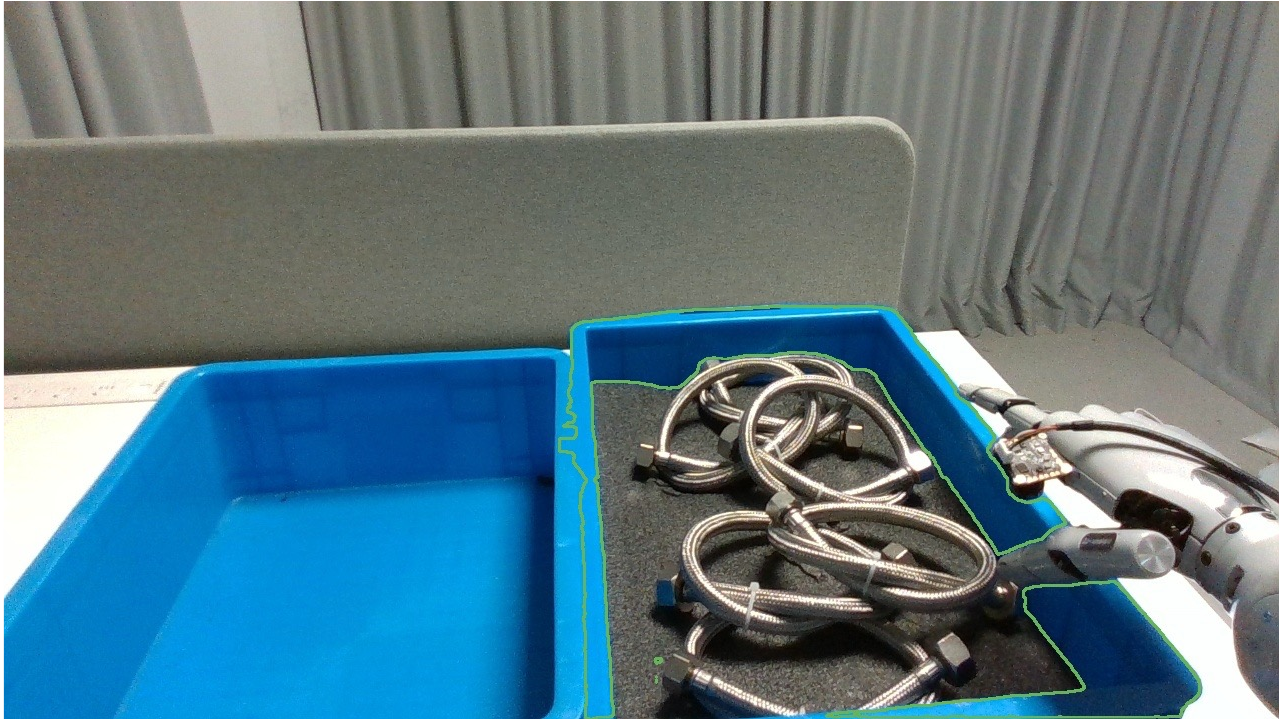}
    \footnotesize (a) Container-region mask.
  \end{minipage}
  \hfill
  \begin{minipage}[t]{0.32\linewidth}
    \centering
    \includegraphics[width=\linewidth]{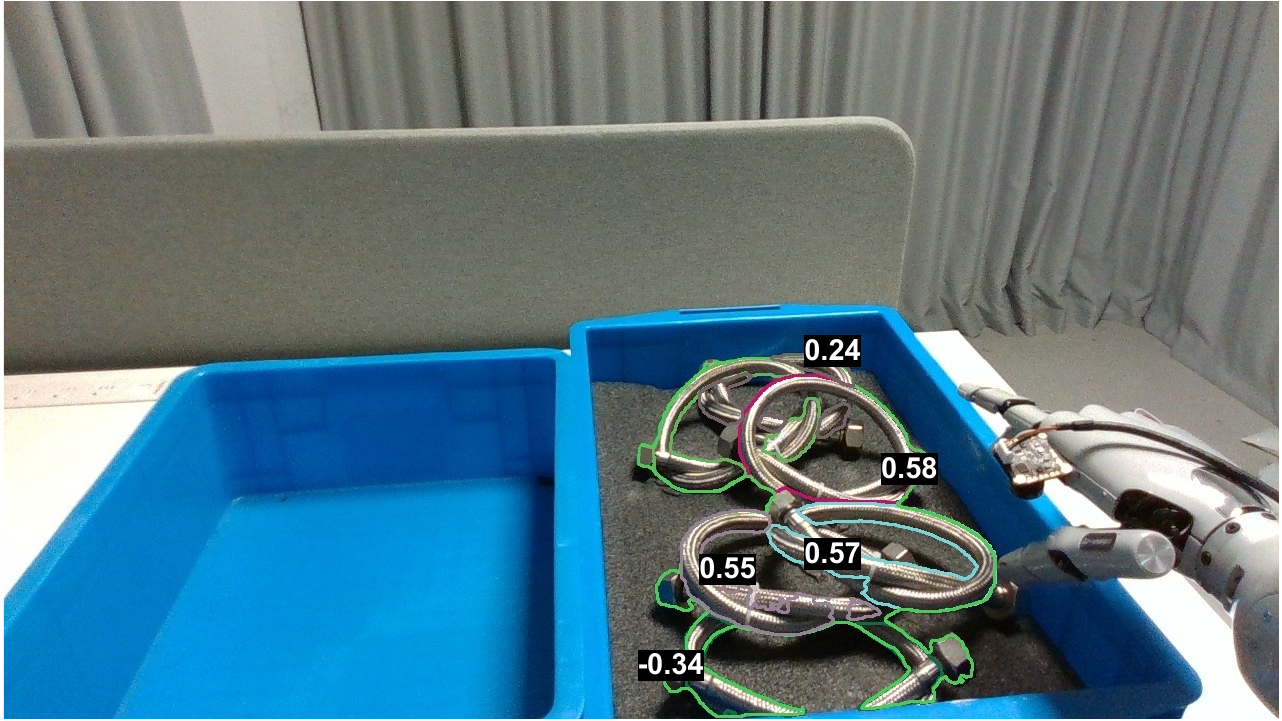}
    \footnotesize (b) Target candidates and scores.
  \end{minipage}
  \hfill
  \begin{minipage}[t]{0.32\linewidth}
    \centering
    \includegraphics[width=\linewidth]{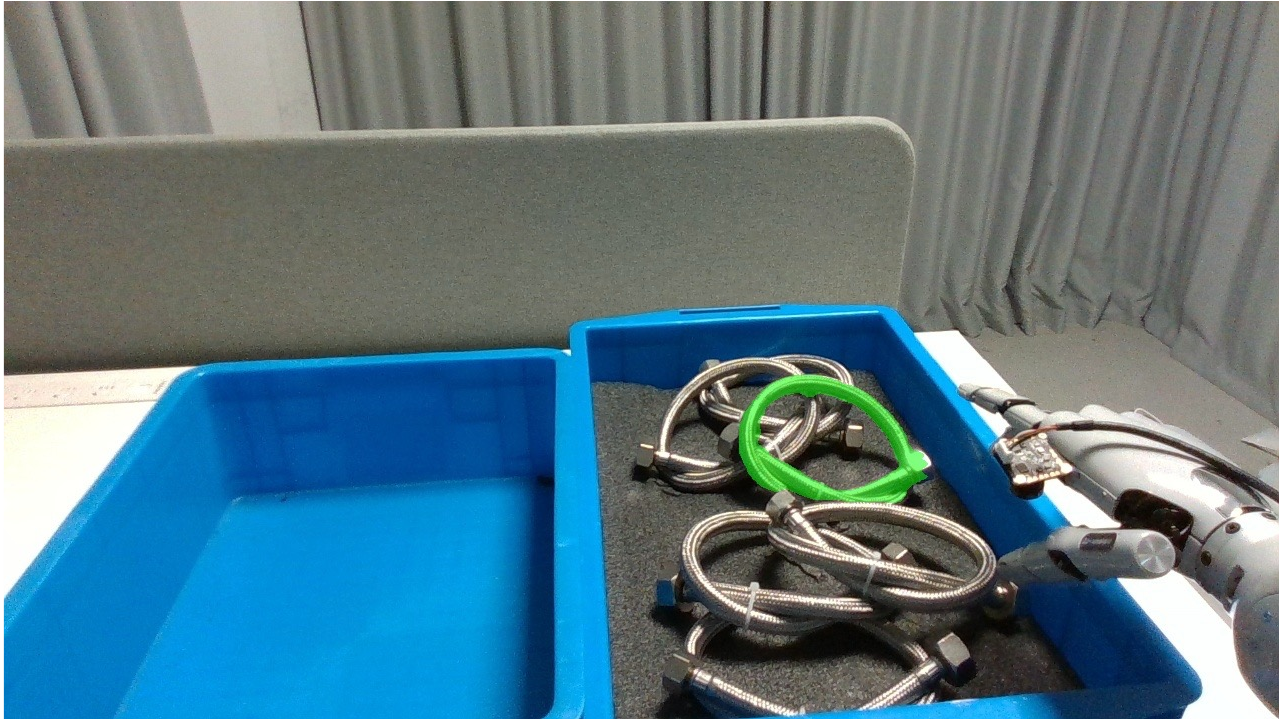}
    \footnotesize (c) Selected overlay.
  \end{minipage}
  \caption{SAM3-based proposal initialization at a subtask boundary. A container prompt first localizes the workspace region, and a target-category prompt generates candidate instance masks with confidence scores. After simple spatial and size filtering, the selected proposal is rendered as the mask overlay used by the policy during closed-loop execution.}
  \label{fig:sam3_pipeline}
\end{figure}

\textbf{Inference initialization.} At each subtask boundary, SAM3 processes the current chest-camera frame using a container text prompt to localize the workspace and a target-category prompt to generate candidate instance masks. We filter candidates using container membership, spatial constraints, and size statistics, then select one proposal using SAM3 confidence scores and these heuristics. Figure~\ref{fig:sam3_pipeline} illustrates this initialization procedure.

\textbf{Overlay rendering.} As noted in Section~\ref{sec:mask_prior}, the binary mask is also rendered as a semi-transparent overlay on the chest-camera image. During training, each trajectory uniformly samples one tint from a fixed palette of vivid RGB colors and alpha-blends the mask with opacity $0.5$; the tint is held fixed within the trajectory to avoid per-frame visual noise. During deployment, each new subtask selects a high-contrast tint with respect to the masked foreground and its surrounding region, and the chosen tint is held fixed until the next subtask transition. The same mask-generation and overlay-rendering pipeline is used wherever a rendered mask prompt is required.

\subsection{Closed-Loop Deployment}
\label{app:deployment}

The selected proposal initializes the SAM3 video tracker, which updates the target mask at each observation. The policy receives the resulting chest RGB overlay together with the wrist RGB view, robot state, and language instruction, and predicts an action chunk and progress $p_t$. When $p_t>0.8$, the current subtask ends and SAM3 is re-initialized to select the next object. Mask tracking operates at the 10\,Hz observation rate ($\leq100$\,ms/frame), while re-segmentation takes approximately 250\,ms and occurs only at subtask transitions. Policy inference latency is evaluated separately in Appendix~\ref{app:efficiency}.

\section{Evaluation Details}
\label{app:evaluation_details}

\subsection{Sample Sizes and Success-Rate Aggregation}
\label{app:stats}

Table~\ref{tab:stats_combined} reports episode and grasp-attempt counts for each method and difficulty split in the in-domain, unseen-object, unseen-lighting, and ablation evaluations. Tables~\ref{tab:main_results}--\ref{tab:ablation} report mean per-episode success rates and standard errors. For unseen objects, we average the means of the two held-out categories, each evaluated over five episodes; the other settings average directly over episodes. Total attempt counts vary because episodes end when the bin is cleared or the $4N$-attempt budget is reached.

\begin{table}[ht]
\centering
\caption{Episode and grasp-attempt counts for the in-domain, unseen-object, unseen-lighting, and ablation evaluations. $\#\mathrm{ep}$ denotes episodes; $\#\mathrm{att}$ denotes grasp attempts.}
\label{tab:stats_combined}
\small
\renewcommand{\arraystretch}{1.05}
\setlength{\tabcolsep}{18pt}
\begin{tabular}{@{}l cc cc@{}}
\toprule
& \multicolumn{2}{c}{\textbf{Easy}} & \multicolumn{2}{c}{\textbf{Difficult}} \\
\cmidrule(lr){2-3}\cmidrule(lr){4-5}
Method & $\#$ep & $\#$att & $\#$ep & $\#$att \\
\midrule
\multicolumn{5}{@{}l}{\emph{In-domain}} \\
GR00T N1.5      & 10 & 118 & 10 & 109 \\
$\pi_0$         & 10 & 172 & 10 & 147 \\
$\pi_{0.5}$     & 10 & 178 & 10 & 188 \\
DexGraspVLA     & 10 & 226 & 10 & 204 \\
FOCUS           & 10 &  91 & 10 &  88 \\
\midrule
\multicolumn{5}{@{}l}{\emph{Unseen Objects}} \\
GR00T N1.5      & 10 & 167 & 10 & 174 \\
$\pi_0$         & 10 & 189 & 10 & 157 \\
$\pi_{0.5}$     & 10 & 192 & 10 & 164 \\
DexGraspVLA     & 10 & 220 & 10 & 162 \\
FOCUS           & 10 &  98 & 10 &  72 \\
\midrule
\multicolumn{5}{@{}l}{\emph{Unseen Lighting}} \\
GR00T N1.5      & 10 & 200 & 10 & 180 \\
$\pi_0$         & 10 & 240 & 10 & 240 \\
$\pi_{0.5}$     & 10 & 240 & 10 & 186 \\
DexGraspVLA     & 10 & 226 & 10 & 192 \\
FOCUS           & 10 & 105 & 10 &  74 \\
\midrule
\multicolumn{5}{@{}l}{\emph{Ablation (in-domain)}} \\
GR00T N1.5      & 10 & 118 & 10 & 109 \\
$+$ Mask overlay       & 10 & 115 & 10 & 128 \\
$+$ Attention bias     & 10 &  91 & 10 &  95 \\
$+$ Latent distillation & 10 &  93 & 10 &  88 \\
FOCUS                  & 10 &  91 & 10 &  88 \\
\bottomrule
\end{tabular}
\end{table}
\FloatBarrier

\subsection{Target-to-Background Attention Ratio}
\label{app:tbr_details}

We evaluate object-region grounding on 100 randomly sampled observations from each Easy and Difficult test split. We extract action-to-vision cross-attention from the final DiT cross-attention block at the final denoising step and average it over attention heads and action queries. The tracked target mask is pooled to the visual-token grid to partition target and background tokens; for RGB-only GR00T N1.5, this mask is used only to compute the metric. For each observation, the target-to-background attention ratio is the geometric mean over target-token attention divided by the geometric mean over background-token attention. Table~\ref{tab:tbr} reports the mean across observations, with 95\% confidence intervals obtained by non-parametric bootstrap resampling.

\section{Training and Inference Efficiency}
\label{app:efficiency}

We quantify the computational trade-off between direct attention bias and latent distillation using the matched configurations in Table~\ref{tab:ablation}. Both retain progress prediction but disable progress-to-action conditioning. The direct-bias configuration retains explicit attention routing at inference, whereas the distilled configuration uses the standard inference pathway. Timings are CUDA-synchronized; training excludes data loading and checkpointing, and inference uses fixed, preprocessed inputs.

\begin{table}[ht]
\centering
\caption{Policy-efficiency profiling on A100-80GB GPUs (training: 4 GPUs, BF16, global batch 64; inference: 1 GPU, batch 1, 4 denoising steps, 16-action chunks). Brackets denote 95\% bootstrap confidence intervals for relative time.}
\label{tab:efficiency}
\small
\renewcommand{\arraystretch}{1.08}
\setlength{\tabcolsep}{7pt}
\begin{tabular}{@{}llcc@{}}
\toprule
Stage & Configuration & Latency $\downarrow$ & Relative time $\downarrow$ \\
\midrule
\multirow{2}{*}{\shortstack[l]{Training\\(ms/step)}}
 & Direct attention bias & \textbf{703.86} & $1.00\times$ \\
 & Latent distillation & 716.99 & $+1.87\%\;[1.65, 2.08]$ \\
\midrule
\multirow{2}{*}{\shortstack[l]{Inference\\(ms/chunk)}}
 & Direct attention bias & 104.21 & $+45.89\%\;[45.80, 45.99]$ \\
 & Latent distillation & \textbf{71.43} & $1.00\times$ \\
\bottomrule
\end{tabular}
\end{table}
\FloatBarrier

Distillation adds 1.87\% to training time, whereas retaining direct attention bias at inference increases policy latency by 45.89\%. These results support distilling attention routing during training rather than retaining it during deployment.

\section{Robustness to Mask Perturbations}
\label{app:mask_perturbations}

To evaluate sensitivity to incomplete masks, we apply two controlled, strictly subtractive perturbations to the SAM3 mask at inference: one iteration of erosion with a $3\!\times\!3$ elliptical kernel and a centered 25\% truncation along the longest dimension of the foreground bounding box. The policy checkpoint and closed-loop protocol remain unchanged.

\begin{table}[ht]
\centering
\caption{FOCUS under controlled mask perturbations (mean per-episode success rate, \% $\pm$ SE; 10 episodes per difficulty split).}
\label{tab:mask_perturbations}
\small
\renewcommand{\arraystretch}{1.05}
\setlength{\tabcolsep}{8pt}
\begin{tabular}{@{}lccc@{}}
\toprule
Mask input & Easy & Diff. & Avg. \\
\midrule
Clean & \textbf{68.05}$_{\pm3.97}$ & \textbf{66.91}$_{\pm2.62}$ & \textbf{67.48}$_{\pm2.38}$ \\
Erosion & 67.36$_{\pm3.63}$ & 63.00$_{\pm3.05}$ & 65.18$_{\pm2.36}$ \\
25\% truncation & 67.24$_{\pm3.10}$ & 65.17$_{\pm3.08}$ & 66.20$_{\pm2.14}$ \\
\bottomrule
\end{tabular}
\end{table}

FOCUS retains 65.18\% and 66.20\% average success under erosion and truncation, respectively, demonstrating robustness to these incomplete-mask perturbations.

\section{Progress Prediction under Pauses and Re-grasping}
\label{app:progress_events}

We construct a 10\,Hz test set of 100 Easy and 100 Difficult teleoperated trajectories containing pauses and re-grasp attempts. Pause, grasp-miss, and retry timestamps are manually annotated, yielding 50 instances of each event type per difficulty split. For each event, the checkpoint trained for that split independently predicts progress from the timestamped observation and from an observation one execution interval later. We define $\Delta p=p_{t+H}-p_t$ and count a prediction as correct when $|\Delta p|\leq0.05$ for a pause, $\Delta p\leq0$ after a missed grasp, and $\Delta p>0$ on a retry. Table~\ref{tab:progress_events} reports event-level accuracy for each event type and difficulty split.

\begin{table}[ht]
\centering
\caption{Event-direction accuracy (\%; 50 annotated instances per event type and difficulty split).}
\label{tab:progress_events}
\small
\renewcommand{\arraystretch}{1.05}
\setlength{\tabcolsep}{10pt}
\begin{tabular}{@{}lccc@{}}
\toprule
Event criterion & Easy & Diff. & Avg. \\
\midrule
Paused ($|\Delta p|\leq0.05$) & 100 & 100 & 100 \\
Grasp miss ($\Delta p\leq0$) & 92 & 72 & 82 \\
Retry ($\Delta p>0$) & 100 & 90 & 95 \\
\bottomrule
\end{tabular}
\end{table}

Table~\ref{tab:progress_events} shows stable progress predictions during pauses and event-conditioned changes following missed grasps and retries. These results indicate sensitivity to execution state rather than elapsed time alone.

\section{Failure Mode Analysis}
\label{app:failure_modes}

We manually inspect all failed FOCUS grasp attempts from the in-domain evaluation in Table~\ref{tab:main_results} and assign each to one of four mutually exclusive categories, ordered by frequency.

\paragraph{Slippage during lift (39.3\%).} The hand makes initial contact with the target object, but the object slips before transfer. Observed cases include slippage on smooth surfaces and incomplete finger closure, where only a subset of fingers stabilizes the object.

\paragraph{Boundary-region grasping (36.1\%).} The hand approaches the seam between touching instances, resulting in an empty grasp or unintended multi-object retrieval. These failures highlight the distinction between identifying the target object region and selecting a feasible contact region.

\paragraph{Height misalignment (21.3\%).} The end effector closes above the target or collides with the bin floor before closure. Such errors may reflect depth ambiguity under heavy clutter.

\paragraph{Subtask-transition errors (3.3\%).} Inappropriate progress-triggered transitions lead to premature or delayed selection of the next object proposal.

Most recorded failures concern grasp execution, particularly contact stability, contact-region selection, and height precision. Subtask-transition errors account for a smaller fraction. This breakdown complements the limitations and potential extensions discussed in Section~\ref{sec:limitations}.

\end{document}